\pdfoutput=1
\documentclass[11pt]{article}
\usepackage[final]{acl}
\usepackage{times}
\usepackage{latexsym}
\usepackage[T1]{fontenc}
\usepackage[utf8]{inputenc}
\usepackage{microtype}
\usepackage{inconsolata}
\usepackage{graphicx}
\usepackage{amsmath}
\usepackage{amsfonts}
\usepackage{multirow}
\usepackage{algorithm}
\usepackage{algorithmic}

\title{DL-QAT: Weight-Decomposed Low-Rank Quantization-Aware Training for Large Language Models}

\author{Wenjin Ke, Zhe Li, Dong Li, Lu Tian, Emad Barsoum \\
  Advanced Micro Devices, Inc., Beijing, China \\
  \texttt{\{wenjing.ke, z.li, d.li, lu.tian, emad.barsoum\}@amd.com} \\
  }

\begin{document}
\maketitle
\begin{abstract}
Improving the efficiency of inference in Large Language Models (LLMs) is a critical area of research. Post-training Quantization (PTQ) is a popular technique, but it often faces challenges at low-bit levels, particularly in downstream tasks. Quantization-aware Training (QAT) can alleviate this problem, but it requires significantly more computational resources. To tackle this, we introduced Weight-\textbf{D}ecomposed \textbf{L}ow-Rank \textbf{Q}uantization-\textbf{A}ware \textbf{T}raining (DL-QAT), which merges the advantages of QAT while training only less than $1\%$ of the total parameters. Specifically, we introduce a group-specific quantization magnitude to adjust the overall scale of each quantization group. Within each quantization group, we use LoRA matrices to update the weight size and direction in the quantization space. We validated the effectiveness of our method on the LLaMA and LLaMA2 model families. The results show significant improvements over our baseline method across different quantization granularities. For instance, for LLaMA-7B, our approach outperforms the previous state-of-the-art method by $4.2\%$ in MMLU on 3-bit LLaMA-7B model. Additionally, our quantization results on pre-trained models also surpass previous QAT methods, demonstrating the superior performance and efficiency of our approach.
\end{abstract}

\section{Introduction}
Large language models (LLMs) have demonstrated exceptional performance across a variety of natural language processing (NLP) tasks. With the growing deployment and use of these models, quantization has become an essential method for reducing memory usage and enhancing computational efficiency. In LLM compression, a range of post-training quantization (PTQ) techniques have been developed, such as weight-only and weight-activation quantization. These techniques generally use a small calibration dataset and apply learning or optimization strategies to quickly transform a pre-trained floating-point model into a quantized version. However, PTQ methods struggle in low-bit quantization, especially in the downstream tasks. Despite the potential benefits, the development of quantization-aware training (QAT) algorithms has been constrained. This is primarily due to the significant data and computational resources required for comprehensive model fine-tuning, making it a costly endeavor.
To address the high computational expense associated with training LLMs, the Parameter-Efficient Fine-Tuning (PEFT) methodology has been introduced. PEFT entails fine-tuning only a fraction of the model's parameters, as opposed to the entirety, thereby enabling the efficient adaptation of pre-trained models to a diverse range of downstream applications. Notably, the Low-Rank Adaptation (LoRA) \cite{hu2021lora} technique, which represents the current state-of-the-art in PEFT, has been shown to achieve performance on par with fully fine-tuned models across various downstream tasks, without necessitating alterations to the model's inference architecture. The conventional approach to generating a quantized model for downstream tasks involves a two-step process: first, the floating-point model is fine-tuned on the downstream tasks; second, PTQ is applied to the fine-tuned model. However, this methodology is not without its drawbacks, as it can be cumbersome and may result in a substantial loss of accuracy. Conversely, directly employing QAT methods can lead to prohibitively high computational costs due to the requirement of end-to-end fine-tuning of all the model's parameters. The objective of our research is to devise a seamless, end-to-end process that yields a quantized model with parameter-efficient fine-tuning, thereby mitigating the aforementioned challenges and enhancing the overall efficiency and effectiveness of model adaptation for downstream tasks.

Building upon these considerations, we propose Weight-\textbf{D}ecomposed \textbf{L}ow-Rank \textbf{Q}uantization-\textbf{A}ware \textbf{T}raining (DL-QAT), a novel end-to-end method designed to enhance the efficiency and effectiveness of model quantization for downstream tasks. DL-QAT decomposes the optimization of quantized weights into two processes: group-specific magnitude training and weight fine-tuning within a predefined quantization space. By incorporating a magnitude term, we calibrate the overall scale for each quantization group, ensuring a more precise representation of the model's parameters. Furthermore, we leverage low-rank matrices $A$ and $B$ to refine the quantized weights, thereby enhancing the model's adaptability to the specific requirements of the downstream tasks. To validate the efficacy of our approach, we conducted comprehensive experiments on the LLaMA and LLaMA2 model families. The results demonstrate a significant improvement over the baseline method, QA-LoRA \cite{xu2023qalora}, across various quantization granularities. Specifically, our method surpasses QA-LoRA by $+4.2\%$ on the MMLU benchmark \cite{hendrycks2020mmlu} and by $+5.5\%$ on the LM-Eval benchmark \cite{eval-harness}. Additionally, when compared to the previous state-of-the-art LLM-QAT method \cite{liu2023llmqat}, our approach achieves lower perplexity on the WikiText-2 dataset \cite{merity2016wikitext2ppl} and higher accuracy on the LM-Eval benchmark, underscoring the superior performance of DL-QAT. LLM-QAT requires fine-tuning the entire model parameters, while we only need to fine-tune less than $1\%$ of the parameters to achieve better results. These findings not only highlight the effectiveness of DL-QAT in achieving competitive accuracy levels but also emphasize its efficiency in terms of both parameters and memory usage. By requiring minimal parameter modifications, DL-QAT offers a compelling alternative to traditional quantization methods, particularly for scenarios where computational resources are limited or where the need for rapid model adaptation is paramount.

\section{Related work}
\textbf{Parameter-Efficient Fine-Tuning.}
LoRA (Low-Rank Adaptation) is a key method in Parameter-Efficient Fine-Tuning (PEFT), training a small number of parameters without altering the model inference process. To enhance its capabilities, variants like AdaLoRA\cite{zhang2023adalora} and Pissa\cite{meng2024pissa} enhance rank via Singular Value Decomposition (SVD), while PLoRA\cite{meng2024periodiclora} accumulates low-rank updates progressively. Further, studies like \cite{zhu2024asymmetry} and LoRA+ \cite{hayou2024lora+} delve into the update mechanisms of LoRA’s $A$ and $B$ matrices. DoRA \cite{liu2024dora} proposed a new optimization approach for LoRA, which decomposes LoRA updates into separate magnitude and direction updates to improve accuracy. Inspired by this idea, we further decompose LoRA quantization-aware training into fine-tuning the magnitude for quantization groups and fine-tuning the weights within the quantization space.

\textbf{Quantization of LLM.} Quantization has been widely used in LLM. Based on whether training is required, quantization can be classified into Post-Training Quantization(PTQ) and Quantization-Aware Training(QAT). PTQ methods requires only a small amount of calibration data to update the quantized weights. For instance, GPTQ ~\cite{frantar2022gptq} utilizes merely 128 data samples to approximate second-order information and achieve the quantized weight. As outliers are crucial for LLM, considerable research is dedicated to addressing outlier issues. SmoothQuant ~\cite{xiao2023smoothquant} effectively shifts the quantization challenge from activations to weights through a mathematically equivalent transformation. QuaRot ~\cite{ashkboos2024quarot} employs Hadamard transformations on the weight matrices and attention modules to mitigate outlier effects. Compared with PTQ methods, QAT methods require more training data and resources, but generally achieve better performance. LLM-QAT ~\cite{liu2023llmqat} leverages data generated by pre-trained LLMs and achieves better performance compared with GPTQ. However, LLM-QAT requires significant training resources.

\textbf{Methods combining LoRA and quantization.}
Building upon LoRA, QLoRA\cite{dettmers2024qlora} was the first to propose a memory-efficient fine-tuning method by quantizing the pretrained model to low-bit and fine-tuning a high-precision LoRA component. This approach enables effective fine-tuning of LLMs within limited memory resources. Subsequent methods such as LoFTQ \cite{li2023loftq} and LQ-LoRA \cite{guo2023lqlora} further optimized the initialization of the LoRA component and reduced the memory required for the quantized pretrained model. However, the combination of a low-bit pretrained model and a high-precision LoRA component still resulted in a high-precision weight after merging, which did not improve inference speed.
To address this issue, QA-LoRA\cite{xu2023qalora} made further improvements on QLoRA by learning an additional high-precision group-wise bias for the quantized model, effectively reducing both time and memory consumption without compromising accuracy. However, QA-LoRA could only perform group-wise fine-tuning, resulting in significant accuracy degradation when the quantization granularity increased.

\section{Methodology}
\subsection{Low-Rank Adaptation and Quantization}
In large language models (LLMs), a linear layer is denoted by $ Y = W \cdot X $, where $W$ represents the weight matrix with dimensions $\mathbb{R}^{C_{out} \times C_{in}}$ and $X$ is the input with dimensions $\mathbb{R}^{C_{in} \times T}$. Here, $C_{out}$ and $C_{in}$ denote the output channel and input channel, respectively, and $T$ represents the sequence length. LoRA (Low-Rank Adaptation) refines the model by introducing two low-rank matrices, \( A \) and \( B \), where \( A \in \mathbb{R}^{r \times C_{in}} \) and \( B \in \mathbb{R}^{C_{out} \times r}\), with \( r \) being the rank of LoRA matrix and \( r \ll C_{in}, C_{out} \). The weight matrix \( W \) is then modified as:
\begin{equation}
  \label{eq:lora}
W = W_{0} + \alpha BA
\end{equation}
where \( W_0 \) represents the pretrained weight matrix that remains frozen during training, and \( \alpha \) is a scaling factor that adjusts the influence of the low-rank adaptation.

For a given bit level $n$, the asymmetric weight quantization and dequantization processes can be described by a specific formula:

\begin{equation}
  \label{eq:qw1}
\tilde{w} = clip\left( \left\lfloor \frac{W - b}{s}  \right\rceil , -2^{n-1}, 2^{n-1}-1 \right)
\end{equation}
\begin{equation}
\label{eq:qw2}
W_{q} = s* \tilde{w} + b
\end{equation}
where \( \tilde{w} \) represents the quantized value, while \( W \) is the original floating-point weight. The scale \( s \) determines the step size between quantization levels, and \( b \) is the offset applied to the weight  before scaling. The round function is denoted by \( \left\lfloor \cdot \right\rceil \), and the $clip$ function ensures that the quantized values stay within the range \( (-2^{n-1}, 2^{n-1} - 1) \). Dequantization involves converting the quantized values back to floating-point weights by scaling the quantized value with \( s \) and adding the offset \( b \), thus retrieving the original weight.

Quantization-Aware Training (QAT) simulates quantization during the forward pass by substituting \( W \) with \( W_{q} \), as depicted in equations \ref{eq:qw1} and \ref{eq:qw2}, and employs the Straight-Through Estimator (STE) for gradient backpropagation to achieve the quantization effect. In LoRA, rather than updating the weight matrix \( W \) directly, the updates are applied to the LoRA matrices \( A \) and \( B \). As a result, the quantization and de-quantization formula is modified accordingly:


\begin{multline}
\label{eq:qw3}
\tilde{w_{'}} = \text{clip}\left( \left\lfloor \frac{W_{0} + \alpha BA - b}{s} \right\rceil, \right. \\
\left. -2^{n-1}, 2^{n-1} - 1 \right)
\end{multline}

\begin{equation}
\label{eq:qw4}
W^{'}_{q} = s * \tilde{w_{'}} + b
\end{equation}

These formulas guarantee the integration of quantization effects into the LoRA weight updates, enabling efficient and precise training with quantization.

\subsection{Weight-Decomposed Quantization}
Rather than directly substituting $W$ with $W^{'}_{q}$ in the quantization formula as indicated in equation \ref{eq:qw4} for QAT, or updating the $A$ and $B$ matrices along with the quantization parameters $s$ and $b$, we separate the joint training of LoRA and quantization into two parts: (1) group-specific magnitude training; (2) weight fine-tuning in the pre-defined quantization space. The quantization process is thus reformulated as follows:
\begin{equation}
  \label{eq:qw5}
  \begin{split}
      W_{q} &= m *  W^{'}_{q} \\
      &= m * (W_{0}+\alpha BA)_{q} \\
        &= m *  (s* \widetilde{(W_{0}+\alpha BA)} + b)
  \end{split}
\end{equation}

Here, $m$ represents a newly introduced hyper-parameter denoting the group-specific magnitude, which matches the number of quantization groups and is identical in size to $s$. The matrix $m$ is initialized as a matrix of all ones. LoRA matrix \(A\) is initialized with a random Gaussian distribution, and \( B \) is initialized as a zero matrix.  The variables \( s \) and \( b \) are initialized to map the range \( (Min(W_{0}), Max(W_{0})) \) to the endpoints of the quantization interval. Therefore, $s_{\text{init}} = \frac{Max-Min}{2^n - 1}$, and $b_{\text{init}} = \frac{{2^{(n-1)} \cdot Max} + (2^{(n-1)} - 1) \cdot Min }{2^n - 1}$.

During the initial training phase, the scale factors $s$ and the biases $b$ are trained to ensure that the quantization updates commence from a well-established quantization space. Specifically, updates are applied only to $s$ and $b$ to obtain their initial values $s_{0}$ and $b_{0}$, which are then frozen. Subsequent training involves parameter optimization in two parts: \textbf{group-specific magnitude training} and \textbf{weight finetuning within the predefined quantization space}. The first part involves adjusting the magnitude term $m$ to set the scale for each quantization group, while in the second part, the $A$ and $B$ matrices are fine-tuned, permitting updates to the quantized weights within the established quantization space.

Our proposed method, DL-QAT, ensures a harmonious balance between the constraints imposed by quantization and the optimization of weights to achieve optimal model performance. By integrating the efficient fine-tuning capabilities of LoRA, DL-QAT not only streamlines the training process but also significantly reduces the associated computational costs and resource expenditure. This synergistic approach allows for the realization of state-of-the-art results while maintaining a high degree of efficiency, making it a compelling choice for scenarios where both performance and resource constraints are of paramount importance.

    

\section{Experiments}

In this section, we assess our approach using both language generation and zero-shot few-shot tasks with open-source models LLaMA-7B/13B \cite{touvron2023llama} and LLaMA2-7B/13B \cite{touvron2023llama2} to demonstrate its effectiveness.

\begin{table*}
\centering
\scalebox{0.75}
{\begin{tabular}{lcccccccccccc}
\hline
\multirow{2}{*}{\textbf{LLaMA}} & \multirow{2}{*}{\textbf{Method}} & \multirow{2}{*}{\textbf{Bits}} & \multicolumn{2}{c}{\textbf{MMLU}} & \multicolumn{8}{c}{\textbf{Common Sense Zero-Shot}} \\ 
 &  & & \textbf{0-Shot} & \textbf{5-Shot} & \textbf{ARC\_C} & \textbf{ARC\_E} & \textbf{BoolQ} & \textbf{HellaSwag} & \textbf{OBQA} & \textbf{PIQA} & \textbf{Winogrande} & \textbf{Avg} \\ \hline
\multirow{6}{*}{\textbf{1-7B}} 

 & - & 16 & 32.1 & 34.6 & 38.2 & 67.3 & 72.9 & 56.3 & 28.4 & 78.2 & 67.1 & 58.3 \\  \cline{2-13}
 & QA-LoRA* & 4 & 37.9 & 38.5 & 44.0 & 71.6 & 75.9 & 57.1 & 30.8 & 78.9 & 67.2 & 60.8 \\ 
 & Ours & 4 & \textbf{40.5} & \textbf{39.9} & \textbf{45.0} & \textbf{75.5} & \textbf{79.8} & \textbf{57.9} & \textbf{36.2} & \textbf{78.9} & \textbf{70.2} & \textbf{63.4} \\ \cline{2-13}
 & QA-LoRA* & 3 & 32.2 & 32.9 & \textbf{41.7} & 71.6 & 76.9 & 54.6 & 28.0 & 77.6 & 64.9 & 59.3 \\ 
 & Ours & 3 & \textbf{36.4} & \textbf{33.9} & 41.0 & \textbf{73.4} & \textbf{78.2} & \textbf{55.3} & \textbf{34.2} & \textbf{78.2} & \textbf{67.5} & \textbf{61.1} \\ \hline
\multirow{6}{*}{\textbf{2-7B}} 
 & - & 16 & 40.7 & 45.5 & 39.9 & 69.3 & 71.1 & 56.7 & 31.8 & 78.3 & 67.1 & 59.2 \\  \cline{2-13}
 
 & QA-LoRA* & 4 & 42.5 & 44.8 & 42.7 & 71.9 & 77.6 & 56.9 & 32.6 & \textbf{79.2} & 68.3 & 61.3 \\ 
 & Ours & 4 & \textbf{44.6} & \textbf{45.0} & \textbf{47.2} & \textbf{77.8} & \textbf{79.3} & \textbf{58.1} & \textbf{35.6} & 78.5 & \textbf{68.5} & \textbf{63.6} \\ \cline{2-13}
 & QA-LoRA* & 3 & 37.9 & 37.9 &  38.1 & 66.6 & 75.0 & 54.0 & 32.0 & 76.0 & 66.5 & 58.3 \\ 
 & Ours & 3 & \textbf{40.5} & \textbf{39.4} & \textbf{41.2} & \textbf{74.4} & \textbf{78.0} & \textbf{54.7} & \textbf{32.2} & \textbf{77.5} & \textbf{68.8} & \textbf{60.9} \\ \hline
\end{tabular}}
\caption{Results of weight-only group-wise quantization with group\_size=128 on LLaMA-7B and LLaMA2-7B. The evaluation includes results for MMLU (both 0-shot and 5-shot settings) and Common Sense QA Zero-shot tasks ($acc$ is reported to maintain consistency with QA-LoRA). * indicates reproduced results. }
\label{exp:group-wise}
\end{table*}

\begin{table*}[h]
\centering
\scalebox{0.78}{

\begin{tabular}{lcccccccccc}
\hline
\multirow{2}{*}{\textbf{LLaMA}} & \multirow{2}{*}{\textbf{Method}} & \multirow{2}{*}{\textbf{W-A-KV}} & \textbf{Wikitext2} & \multicolumn{7}{c}{\textbf{Common Sense Zero-Shot}} \\ 
&  &  & \textbf{ppl} ($\downarrow$)  & \textbf{ARC\_C} & \textbf{ARC\_E} & \textbf{BoolQ} & \textbf{HellaSwag} & \textbf{PIQA} & \textbf{Winogrande} & \textbf{Avg} \\
\hline
\multirow{5}{*}{1-7B} 
 & - & 16-16-16 & 5.68 & 48.0 & 73.0 & 76.8 & 76.1  & 79.3 & 70.0 & 70.5 \\ \cline{2-11}
& QA-LoRA* & 3-16-16 & 16.5 & 38.4 & 51.5 & 64.3 & 64.5  & 73.7 & 60.9 & 58.9 \\
& Ours & 3-16-16 & \textbf{9.2} & \textbf{40.1} & \textbf{61.8} & \textbf{71.2} & \textbf{67.2}  & \textbf{75.9} & \textbf{64.0} & \textbf{63.4} \\ \cline{2-11}
& QA-LoRA* & 4-16-16 & 11.1 & 42.4 & 58.0 & 73.7 & 70.5  & 77.3 & 66.1 & 64.7 \\
& LLM-QAT & 4-16-16 & - & \textbf{45.0} & \textbf{70.0} & 75.5 & 74.0  & \textbf{78.3} & \textbf{69.0} & 68.6 \\
& Ours & 4-16-16 & \textbf{6.7} & 44.4 & 68.5 & \textbf{78.5} & \textbf{74.4}  & 78.1 & 68.5 & \textbf{68.7} \\ \cline{2-11}

 & SmoothQuant & 4-8-8 & - & 42.8 & 67.4 & 71.0 & 67.8  & 77.6 & 66.0 & 65.2 \\
 & LLM-QAT & 4-8-8 & - & 45.6 & 70.2 & 74.6 & 73.5  & 77.5 & 67.7 & 68.2 \\
 & Ours & 4-8-8 & \textbf{6.7} & \textbf{46.2} & \textbf{71.3} & \textbf{78.1} & \textbf{73.6}  & \textbf{78.5} & \textbf{68.4} & \textbf{69.4} \\
\hline
\multirow{3}{*}{1-13B} 
& - & 16-16-16 & 5.09 & 52.6 & 74.5 & 78.1 & 79.2  & 80.0 & 73.6 & 73.0 \\ \cline{2-11}
& SmoothQuant & 4-8-8 & - & 43.3 & 67.4 & 72.5 & 74.3  & 77.1 & 69.5 & 67.4 \\
 & LLM-QAT & 4-8-8 & - & \textbf{51.9} & 73.6 & 77.5 & 73.6  & 79.1 & \textbf{70.6} & 71.6 \\
 & Ours & 4-8-8 & \textbf{5.9} & 48.8 & \textbf{74.8} & \textbf{80.5} & \textbf{77.1}  & \textbf{80.4} & 70.3 & \textbf{72.0} \\
\hline
\multirow{5}{*}{2-7B} 
& - & 16-16-16 & 5.47 & 46.3 & 74.6 & 77.7 & 76.0  & 79.1 & 69.1 & 70.5 \\ \cline{2-11}
 & QA-LoRA* & 3-16-16 & 13.7 & \textbf{36.3} & 48.2 & 70.3 & \textbf{66.3}  & 74.4 & \textbf{63.9} & 59.9 \\
 & Ours & 3-16-16 & \textbf{9.4} & 35.9 & \textbf{58.9} & \textbf{71.1} & 63.6  & \textbf{74.8} & 60.2 & \textbf{63.7} \\ \cline{2-11}
 & QA-LoRA* & 4-16-16 & 9.5 & 41.3 & 55.1 & 68.8 & 71.9  & 77.3 & 68.2 & 63.8 \\
 & Ours & 4-16-16 & \textbf{6.3} & \textbf{44.6} & \textbf{71.0} & \textbf{78.5} & \textbf{74.6}  & \textbf{78.2} & \textbf{68.8} & \textbf{69.3} \\ \cline{2-11}
\hline
2-13B 
& - & 16-16-16 & 4.88 & 49.0 & 77.4 & 80.6 & 79.4  & 80.5 & 72.2 & 73.2 \\ \cline{2-11}
& Ours & 4-8-8 & 5.63 & 49.6 & 75.5 & 81.9 & 78.1  & 80.1 & 70.3 & 72.6 \\
\hline
\end{tabular}
}

\caption{Results of channel-wise quantization results on LLaMA-7B/13B and LLaMA2-7B/13B models. Evaluation metrics include perplexity (ppl) on WikiText-2 and accuracy in common sense QA zero-shot tasks. $Acc\_norm$ is reported to ensure consistency with LLM-QAT. * indicates reproduced results.}
\label{exp:channel-wise}
\end{table*}

\subsection{Experiment Setup} 
\textbf{Dataset.}
We use Stanford-Alpaca dataset \cite{taori2023alpaca} as the fine-tuning dataset. Alpaca comprises a dataset of 52,000 instructions and demonstrations created by OpenAI's text-davinci-003 engine. This instructional data can be utilized to perform instruction-tuning on language models, enhancing their ability to follow instructions more effectively.

\noindent\textbf{Training Details.}
In all experiments, a batch size of 16 was maintained, and a constant learning rate of 2e-4 was used. The optimizer employed was \texttt{adamw\_hf}, with the default LoRA rank set at 16. For consistency with QALoRA's settings, training was conducted for 10,000 iterations, while other experimental results underwent 5,000 iterations. The training iterations for learning \(s_{0}\) and \(b_{0}\) were uniformly set at 1000. This approach ensures fair comparisons and reliable results across various models and datasets. Our experimental setup involves a quantization simulation in which all learnable parameters are represented in bf16 format. During inference, these quantized weights are dequantized back to bf16 for computation. We conducted all experiments on AMD MI-250 GPUs to maintain consistent hardware conditions.

\noindent\textbf{Evaluation Tasks.}
The evaluation encompassed a broad spectrum of benchmarks. For language generation tasks, the perplexity on WikiText-2 \cite{merity2016wikitext2ppl} was reported. Additionally, results on the Massively Multitask Language Understanding (MMLU) benchmark \cite{hendrycks2020mmlu} were presented in both zero-shot and five-shot settings. The method was also assessed on seven common sense reasoning tasks from the EleutherAI LM Harness \cite{eval-harness} for zero-shot performance.

\subsection{Results}
Our evaluation spanned various quantization granularities, including group-wise and channel-wise quantization. In group-wise quantization, we employed a standard setting with a group size of 128. For channel-wise quantization, our experiments encompassed two scenarios: one with quantization applied solely to weights, and another with quantization extended to weights, activations, and the kv cache.

Our approach was evaluated against prior quantization-aware LoRA-based methods, using QA-LoRA as the benchmark. To ensure a thorough comparison, we replicated the QA-LoRA algorithm with a group size of 128 and channel-wise quantization, while preserving its original LoRA rank of 64. The results presented in Table \ref{exp:group-wise} and Table \ref{exp:channel-wise} demonstrate that our technique surpasses the benchmark across different quantization bits, granularities, and datasets. Remarkably, we noted a $+4.2\%$ enhancement in MMLU zero-shot accuracy on LLaMA-7B with 3-bit group-wise quantization, and a $+5.5\%$ increase in Common Sense QA accuracy on LLaMA2-7B with 4-bit per-channel quantization.

Moreover, we conducted comparisons with the PTQ method SmoothQuant \cite{xiao2023smoothquant} and the QAT method LLM-QAT \cite{liu2023llmqat} on the LLaMA-7B/13B models within the W4A8KV8 framework, as depicted in Table \ref{exp:channel-wise}. Our approach yielded lower perplexity scores compared to LLM-QAT. In terms of common sense QA accuracy, it substantially surpasses SmoothQuant and LLM-QAT. Moreover, our approach necessitates significantly less training memory and time compared to LLM-QAT, proving that our DL-QAT method not only yields superior outcomes but also enhances efficiency.

\begin{table*}
\centering
\scalebox{0.9}
{\begin{tabular}{llllll}
\hline
\multirow{2}{*}{\textbf{Setting}} & \multirow{2}{*}{\textbf{m}} & \multirow{2}{*}{\textbf{Clipping bounds}} & \multirow{2}{*}{\textbf{Learnable params}} & \multicolumn{2}{c}{\textbf{LLaMA-7B}}  \\
&&&& \textbf{4 bit} & \textbf{3 bit} \\
\hline
1 & N/A & MinMax & $A,B$ & 69.7 & 67.5 \\
2 & N/A & Learn then fix & $s,b$ then $A,B$ & 70.4 & 66.7 \\
3 & N/A & Learn & $s,b,A,B$ &70.0 & 67.2 \\
\hline
4 & Learn & MinMax  & $m, A,B$ & 70.4 & 67.2 \\
5 & Learn & Learn then fix & $s,b$ then $m, A,B$ & \textbf{70.7} & \textbf{68.3} \\
6 & Learn & Learn & $m,s,b,A,B$ & 70.0 & 67.7 \\
\hline
\end{tabular}}
\caption{Results with different magnitude and quantization settings on LLaMA-7B. Average $acc\_norm$ in common sense QA zero-shot tasks is reported. With a quantization granularity of group\_size=128. }
\label{tab:Q setting ab}

\end{table*}

\begin{table*}
\centering
\scalebox{0.9}
{\begin{tabular}{llcccc}
\hline
 \multirow{2}{*}{\textbf{LLaMA}} & \multirow{2}{*}{\textbf{Quant config}} & \multicolumn{2}{l}{\textbf{Trainable Params (M)}} & \multirow{2}{*}{\textbf{GPU Memory (G)}} &\multirow{2}{*} {\textbf{Training speed (s/iter)}} \\ 
& & $s,b$ & $m,A,B$ && \\
\hline
\multirow{3}{*}{\textbf{7B}} 
& Weight-only, g128 & 50 & 71 & 32.5 &3.33 \\ 
& Weight-only, per-channel &1&41&31.8 &3.24 \\ 
& Quant W/A/KV, per-channel &1&41&33.1&3.91\\ 
\hline
\multirow{3}{*}{\textbf{13B}} 
& Weight-only, g128 &99&162&60.4&6.26 \\ 
& Weight-only, per-channel &2&65&58.7&6.09 \\ 
& Quant W/A/KV, per-channel &2&65&62.8&7.04 \\
\hline
\end{tabular}}
\caption{Training parameter count, GPU memory usage, and training speed for LLaMA-7B/13B under different quantization configurations  with a per-GPU batch size of 16. The experiments were conducted on an AMD MI250 with 64GB of GPU memory. }
\label{tab:training efficiency}
\end{table*}

\subsection{Ablation Study}
To demonstrate the effectiveness of our introduced group-specific magnitude $m$ and our quantization update strategy, including weight fine-tuning in the pre-defined quantization space, we conducted ablation experiments as shown in Table \ref{tab:Q setting ab}. 

For quantization updates, we considered three possible settings: (1) Min-Max Clipping Values: Quantization values are uniformly distributed between the updated $\text{min}(W_{0} + \alpha BA)$ and $\text{max}(W_{0} + \alpha BA)$, with clipping always performed at these dynamic bounds. (2) Fixed Clipping Values: The clipping values are fixed by learned $s_{0}$ and $b_{0}$, ensuring that $W_{0} + \alpha BA$ updates within a fixed quantization space. (3) Adaptive Clipping Values: Both $s$ and $b$ are continuously trained, adaptively updating the quantization space throughout the training process. For the magnitude $m$, we explored two possible settings: with or without the learnable magnitude term $m$.

The results in Table \ref{tab:Q setting ab} show that experiments with the learnable magnitude $m$ consistently outperform those without it. This indicates that using $m$ to adjust the quantization group's magnitude aids in adaptive scaling. Without the learnable magnitude $m$, accuracy across various bit settings varies, with no single setting being clearly superior. However, when combined with the learnable magnitude $m$, setting 2 — our proposed method of weight fine-tuning in the pre-defined quantization space — significantly outperforms the other settings. This suggests that our strategy of decomposing the weight into two parts for updates is effective, allowing the magnitude and weight distribution to be optimized separately, resulting in excellent fine-tuning outcomes.


\subsection{Analysis}

In Table \ref{tab:training efficiency}, we evaluate the training parameter count, GPU memory usage, and training speed for LLaMA-7B and 13B models. The total parameters of LLaMA-7B and LLaMA-13B are 6.8G and 13.1G, respectively. For group-wise quantization, after fixing parameters $s$ and $b$, the remaining trainable parameters $m$ and $A, B$ account for only $1.0\%$ and $1.2\%$ of the total parameters in LLaMA-7B and LLaMA-13B, respectively. For channel-wise quantization, the training parameters constitute $0.6\%$ and $0.5\%$ of the total parameters for LLaMA-7B and LLaMA-13B, respectively. With a batch size of 16, our simulated quantized training shows that LLaMA 7B and 13B use a maximum of $33.1GB$ and $62.8GB$ of GPU memory, respectively. On the Alpaca dataset, with an AMD MI250 GPU, LLaMA-7B can train up to $17,669$ samples per hour, while LLaMA-13B can train up to $9,458$ samples per hour. Therefore, compared to the previous QAT methods, our approach takes only about one-thirtieth of the time to converge the model, significantly reducing the resources needed for training.

\section{Conclusion}

In this paper, we introduce Weight-\textbf{D}ecomposed \textbf{L}ow-Rank \textbf{Q}uantization-\textbf{A}ware \textbf{T}raining (DL-QAT), a novel end-to-end approach designed to improve the efficiency of QAT for tasks downstream of LLMs. DL-QAT optimizes quantized weights through two main processes: group-specific magnitude training and weight fine-tuning within a set quantization space. By employing Low-Rank Adaptation (LoRA) matrices, we are able to update the weight magnitude and direction within the quantization space, thereby enabling precise adjustments to the model's parameters. DL-QAT achieves remarkable results by training on less than $1\%$ of the model's parameters, outperforming previous QAT methods across established Natural Language Processing benchmarks. This efficiency in parameter utilization is a testament to the effectiveness of DL-QAT in achieving state-of-the-art performance while minimizing computational overhead.

\bibliography{reference}

\begin{thebibliography}{21}
\providecommand{\natexlab}[1]{#1}

\bibitem[{Ashkboos et~al.(2024)Ashkboos, Mohtashami, Croci, Li, Jaggi, Alistarh, Hoefler, and Hensman}]{ashkboos2024quarot}
Saleh Ashkboos, Amirkeivan Mohtashami, Maximilian~L Croci, Bo~Li, Martin Jaggi, Dan Alistarh, Torsten Hoefler, and James Hensman. 2024.
\newblock Quarot: Outlier-free 4-bit inference in rotated llms.
\newblock \emph{arXiv preprint arXiv:2404.00456}.

\bibitem[{Dettmers et~al.(2024)Dettmers, Pagnoni, Holtzman, and Zettlemoyer}]{dettmers2024qlora}
Tim Dettmers, Artidoro Pagnoni, Ari Holtzman, and Luke Zettlemoyer. 2024.
\newblock Qlora: Efficient finetuning of quantized llms.
\newblock \emph{Advances in Neural Information Processing Systems}, 36.

\bibitem[{Frantar et~al.(2022)Frantar, Ashkboos, Hoefler, and Alistarh}]{frantar2022gptq}
Elias Frantar, Saleh Ashkboos, Torsten Hoefler, and Dan Alistarh. 2022.
\newblock Gptq: Accurate post-training quantization for generative pre-trained transformers.
\newblock \emph{arXiv preprint arXiv:2210.17323}.

\bibitem[{Gao et~al.(2023)Gao, Tow, Abbasi, Biderman, Black, DiPofi, Foster, Golding, Hsu, Le~Noac'h, Li, McDonell, Muennighoff, Ociepa, Phang, Reynolds, Schoelkopf, Skowron, Sutawika, Tang, Thite, Wang, Wang, and Zou}]{eval-harness}
Leo Gao, Jonathan Tow, Baber Abbasi, Stella Biderman, Sid Black, Anthony DiPofi, Charles Foster, Laurence Golding, Jeffrey Hsu, Alain Le~Noac'h, Haonan Li, Kyle McDonell, Niklas Muennighoff, Chris Ociepa, Jason Phang, Laria Reynolds, Hailey Schoelkopf, Aviya Skowron, Lintang Sutawika, Eric Tang, Anish Thite, Ben Wang, Kevin Wang, and Andy Zou. 2023.
\newblock \href {https://doi.org/10.5281/zenodo.10256836} {A framework for few-shot language model evaluation}.

\bibitem[{Guo et~al.(2023)Guo, Greengard, Xing, and Kim}]{guo2023lqlora}
Han Guo, Philip Greengard, Eric~P Xing, and Yoon Kim. 2023.
\newblock Lq-lora: Low-rank plus quantized matrix decomposition for efficient language model finetuning.
\newblock \emph{arXiv preprint arXiv:2311.12023}.

\bibitem[{Hayou et~al.(2024)Hayou, Ghosh, and Yu}]{hayou2024lora+}
Soufiane Hayou, Nikhil Ghosh, and Bin Yu. 2024.
\newblock Lora+: Efficient low rank adaptation of large models.
\newblock \emph{arXiv preprint arXiv:2402.12354}.

\bibitem[{Hendrycks et~al.(2020)Hendrycks, Burns, Basart, Zou, Mazeika, Song, and Steinhardt}]{hendrycks2020mmlu}
Dan Hendrycks, Collin Burns, Steven Basart, Andy Zou, Mantas Mazeika, Dawn Song, and Jacob Steinhardt. 2020.
\newblock Measuring massive multitask language understanding.
\newblock \emph{arXiv preprint arXiv:2009.03300}.

\bibitem[{Hu et~al.(2021)Hu, Shen, Wallis, Allen-Zhu, Li, Wang, Wang, and Chen}]{hu2021lora}
Edward~J Hu, Yelong Shen, Phillip Wallis, Zeyuan Allen-Zhu, Yuanzhi Li, Shean Wang, Lu~Wang, and Weizhu Chen. 2021.
\newblock Lora: Low-rank adaptation of large language models.
\newblock \emph{arXiv preprint arXiv:2106.09685}.

\bibitem[{Li et~al.(2023)Li, Yu, Liang, He, Karampatziakis, Chen, and Zhao}]{li2023loftq}
Yixiao Li, Yifan Yu, Chen Liang, Pengcheng He, Nikos Karampatziakis, Weizhu Chen, and Tuo Zhao. 2023.
\newblock Loftq: Lora-fine-tuning-aware quantization for large language models.
\newblock \emph{arXiv preprint arXiv:2310.08659}.

\bibitem[{Liu et~al.(2024)Liu, Wang, Yin, Molchanov, Wang, Cheng, and Chen}]{liu2024dora}
Shih-Yang Liu, Chien-Yi Wang, Hongxu Yin, Pavlo Molchanov, Yu-Chiang~Frank Wang, Kwang-Ting Cheng, and Min-Hung Chen. 2024.
\newblock Dora: Weight-decomposed low-rank adaptation.
\newblock \emph{arXiv preprint arXiv:2402.09353}.

\bibitem[{Liu et~al.(2023)Liu, Oguz, Zhao, Chang, Stock, Mehdad, Shi, Krishnamoorthi, and Chandra}]{liu2023llmqat}
Zechun Liu, Barlas Oguz, Changsheng Zhao, Ernie Chang, Pierre Stock, Yashar Mehdad, Yangyang Shi, Raghuraman Krishnamoorthi, and Vikas Chandra. 2023.
\newblock Llm-qat: Data-free quantization aware training for large language models.
\newblock \emph{arXiv preprint arXiv:2305.17888}.

\bibitem[{Meng et~al.(2024{\natexlab{a}})Meng, Wang, and Zhang}]{meng2024pissa}
Fanxu Meng, Zhaohui Wang, and Muhan Zhang. 2024{\natexlab{a}}.
\newblock Pissa: Principal singular values and singular vectors adaptation of large language models.
\newblock \emph{arXiv preprint arXiv:2404.02948}.

\bibitem[{Meng et~al.(2024{\natexlab{b}})Meng, Dai, Luo, Yang, Wu, Wang, Wang, Dong, Chen, and Sui}]{meng2024periodiclora}
Xiangdi Meng, Damai Dai, Weiyao Luo, Zhe Yang, Shaoxiang Wu, Xiaochen Wang, Peiyi Wang, Qingxiu Dong, Liang Chen, and Zhifang Sui. 2024{\natexlab{b}}.
\newblock Periodiclora: Breaking the low-rank bottleneck in lora optimization.
\newblock \emph{arXiv preprint arXiv:2402.16141}.

\bibitem[{Merity et~al.(2016)Merity, Xiong, Bradbury, and Socher}]{merity2016wikitext2ppl}
Stephen Merity, Caiming Xiong, James Bradbury, and Richard Socher. 2016.
\newblock Pointer sentinel mixture models.
\newblock \emph{arXiv preprint arXiv:1609.07843}.

\bibitem[{Taori et~al.(2023)Taori, Gulrajani, Zhang, Dubois, Li, Guestrin, Liang, and Hashimoto}]{taori2023alpaca}
Rohan Taori, Ishaan Gulrajani, Tianyi Zhang, Yann Dubois, Xuechen Li, Carlos Guestrin, Percy Liang, and Tatsunori~B Hashimoto. 2023.
\newblock Alpaca: A strong, replicable instruction-following model.
\newblock \emph{Stanford Center for Research on Foundation Models. https://crfm. stanford. edu/2023/03/13/alpaca. html}, 3(6):7.

\bibitem[{Touvron et~al.(2023{\natexlab{a}})Touvron, Lavril, Izacard, Martinet, Lachaux, Lacroix, Rozi{\`e}re, Goyal, Hambro, Azhar et~al.}]{touvron2023llama}
Hugo Touvron, Thibaut Lavril, Gautier Izacard, Xavier Martinet, Marie-Anne Lachaux, Timoth{\'e}e Lacroix, Baptiste Rozi{\`e}re, Naman Goyal, Eric Hambro, Faisal Azhar, et~al. 2023{\natexlab{a}}.
\newblock Llama: Open and efficient foundation language models.
\newblock \emph{arXiv preprint arXiv:2302.13971}.

\bibitem[{Touvron et~al.(2023{\natexlab{b}})Touvron, Martin, Stone, Albert, Almahairi, Babaei, Bashlykov, Batra, Bhargava, Bhosale et~al.}]{touvron2023llama2}
Hugo Touvron, Louis Martin, Kevin Stone, Peter Albert, Amjad Almahairi, Yasmine Babaei, Nikolay Bashlykov, Soumya Batra, Prajjwal Bhargava, Shruti Bhosale, et~al. 2023{\natexlab{b}}.
\newblock Llama 2: Open foundation and fine-tuned chat models.
\newblock \emph{arXiv preprint arXiv:2307.09288}.

\bibitem[{Xiao et~al.(2023)Xiao, Lin, Seznec, Wu, Demouth, and Han}]{xiao2023smoothquant}
Guangxuan Xiao, Ji~Lin, Mickael Seznec, Hao Wu, Julien Demouth, and Song Han. 2023.
\newblock Smoothquant: Accurate and efficient post-training quantization for large language models.
\newblock In \emph{International Conference on Machine Learning}, pages 38087--38099. PMLR.

\bibitem[{Xu et~al.(2023)Xu, Xie, Gu, Chen, Chang, Zhang, Chen, Zhang, and Tian}]{xu2023qalora}
Yuhui Xu, Lingxi Xie, Xiaotao Gu, Xin Chen, Heng Chang, Hengheng Zhang, Zhensu Chen, Xiaopeng Zhang, and Qi~Tian. 2023.
\newblock Qa-lora: Quantization-aware low-rank adaptation of large language models.
\newblock \emph{arXiv preprint arXiv:2309.14717}.

\bibitem[{Zhang et~al.(2023)Zhang, Chen, Bukharin, Karampatziakis, He, Cheng, Chen, and Zhao}]{zhang2023adalora}
Qingru Zhang, Minshuo Chen, Alexander Bukharin, Nikos Karampatziakis, Pengcheng He, Yu~Cheng, Weizhu Chen, and Tuo Zhao. 2023.
\newblock Adalora: Adaptive budget allocation for parameter-efficient fine-tuning.
\newblock \emph{arXiv preprint arXiv:2303.10512}.

\bibitem[{Zhu et~al.(2024)Zhu, Greenewald, Nadjahi, Borde, Gabrielsson, Choshen, Ghassemi, Yurochkin, and Solomon}]{zhu2024asymmetry}
Jiacheng Zhu, Kristjan Greenewald, Kimia Nadjahi, Haitz S{\'a}ez de~Oc{\'a}riz Borde, Rickard~Br{\"u}el Gabrielsson, Leshem Choshen, Marzyeh Ghassemi, Mikhail Yurochkin, and Justin Solomon. 2024.
\newblock Asymmetry in low-rank adapters of foundation models.
\newblock \emph{arXiv preprint arXiv:2402.16842}.

\end{thebibliography}




\end{document}